\documentclass[preprint,12pt]{elsarticle}

\usepackage{listings}
\usepackage{color}

\definecolor{dkgreen}{rgb}{0,0.6,0}
\definecolor{gray}{rgb}{0.5,0.5,0.5}
\definecolor{mauve}{rgb}{0.58,0,0.82}

\usepackage{tikz}
\usetikzlibrary{backgrounds}
\usepackage{graphicx}
\usepackage{bm}
\usepackage{amsmath}
\usepackage{setspace}
\usepackage{hyperref}
\usepackage{subfigure}
\usepackage{multirow}
\hypersetup{
    colorlinks=true,       % false: boxed links; true: colored links
}
\usepackage{amssymb}
\usepackage{lineno}
\usepackage{diagbox}

\journal{Journal Name}

\begin{document}

\begin{frontmatter}


\title{Physics Informed Deep Learning (Part I): Data-driven Solutions of Nonlinear Partial Differential Equations}



\author{Maziar Raissi$^{1}$, Paris Perdikaris$^{2}$, and George Em Karniadakis$^{1}$}
\address{$^{1}$Division of Applied Mathematics, Brown University,\\ Providence, RI, 02912, USA\\
$^{2}$Department of Mechanical Engineering and Applied Mechanics,\\ University of Pennsylvania,\\ Philadelphia, PA, 19104, USA}
%\address{Division of Applied Mathematics, Brown University,\\ Providence, RI, 02912, USA}

\begin{abstract}

We introduce {\em physics informed neural networks} -- neural networks that are trained to solve supervised learning tasks while respecting any given law of physics described by general nonlinear partial differential equations. In this two part treatise, we present our developments in the context of solving two main classes of problems: data-driven solution and data-driven discovery of partial differential equations. Depending on the nature and arrangement of the available data, we devise two distinct classes of algorithms, namely continuous time and discrete time models. The resulting neural networks form a new class of {\em data-efficient} universal function approximators that naturally encode any underlying physical laws as prior information. In this first part, we demonstrate how these networks can be used to infer solutions to partial differential equations, and obtain physics-informed surrogate models that are fully differentiable with respect to all input coordinates and free parameters. 

\end{abstract}


\end{frontmatter}

% \linenumbers

\section{Introduction}

With the explosive growth of available data and computing resources, recent advances in machine learning and data analytics have yielded transformative results across diverse scientific disciplines, including image recognition \cite{krizhevsky2012imagenet}, natural language processing \cite{lecun2015deep}, cognitive science \cite{lake2015human}, and genomics \cite{alipanahi2015predicting}. However, more often than not, in the course of analyzing complex physical, biological or engineering systems, the cost of data acquisition is prohibitive, and we are inevitably faced with the challenge of drawing conclusions and making decisions under partial information. In this {\em small data} regime, the vast majority of state-of-the art machine learning techniques (e.g., deep/convolutional/recurrent neural networks) are lacking robustness and fail to provide any guarantees of convergence.\\

At first sight, the task of training a deep learning algorithm to accurately identify a nonlinear map from a few -- potentially very high-dimensional -- input and output data pairs seems at best naive. Coming to our rescue, for many cases pertaining to the modeling of physical and biological systems, there a exist a vast amount of prior knowledge that is currently not being utilized in modern machine learning practice. Let it be the principled physical laws that govern the time-dependent dynamics of a system, or some empirical validated rules or other domain expertise, this prior information can act as a regularization agent that constrains the space of admissible solutions to a manageable size (for e.g., in incompressible fluid dynamics problems by discarding any non realistic flow solutions that violate the conservation of mass principle). In return, encoding such structured information into a learning algorithm results in amplifying the information content of the data that the algorithm sees, enabling it to quickly steer itself towards the right solution and generalize well even when only a few training examples are available.\\

The first glimpses of promise for exploiting structured prior information to construct data-efficient and physics-informed learning machines have already been showcased in the recent studies of \cite{raissi2017inferring, raissi2017machine, owhadi2015bayesian}. There, the authors employed Gaussian process regression \cite{Rasmussen06gaussianprocesses} to devise functional representations that are tailored to a given linear operator, and were able to accurately infer solutions and provide uncertainty estimates for several prototype problems in mathematical physics. Extensions to nonlinear problems were proposed in subsequent studies by Raissi {\em et. al.} \cite{raissi2017numerical, raissi2017hidden} in the context of both inference and systems identification. Despite the flexibility and mathematical elegance of Gaussian processes in encoding prior information, the treatment of nonlinear problems introduces two important limitations. First, in \cite{raissi2017numerical,raissi2017hidden} the authors had to locally linearize any nonlinear terms in time, thus limiting the applicability of the proposed methods to discrete-time domains and compromising the accuracy of their predictions in strongly nonlinear regimes. Secondly, the Bayesian nature of Gaussian process regression requires certain prior assumptions that may limit the representation capacity of the model and give rise to robustness/brittleness issues, especially for nonlinear problems \cite{owhadi2015brittleness}.

\subsection{Problem setup and summary of contributions}
In this work we take a different approach by employing deep neural networks and leverage their well known capability as universal function approximators \cite{hornik1989multilayer}. In this setting, we can directly tackle nonlinear problems without the need for committing to any prior assumptions, linearization, or local time-stepping. We exploit recent developments in automatic differentiation \cite{baydin2015automatic} -- one of the most useful but perhaps underused techniques in scientific computing -- to differentiate neural networks with respect to their input coordinates and model parameters to obtain {\em physics informed neural networks}. Such neural networks are constrained to respect any symmetry, invariance, or conservation principles originating from the physical laws that govern the observed data, as modeled by general time-dependent and nonlinear partial differential equations. This simple yet powerful construction allows us to tackle a wide range of problems in computational science and introduces a potentially disruptive technology leading to the development of new data-efficient and physics-informed learning machines, new classes of numerical solvers for partial differential equations, as well as new data-driven approaches for model inversion and systems identification.\\

The general aim of this work is to set the foundations for a new paradigm in modeling and computation that enriches deep learning with the longstanding developments in mathematical physics. These developments are presented in the context of two main problem classes: data-driven solution and data-driven discovery of partial differential equations. To this end, let us consider parametrized and nonlinear partial differential equations of the general form
\begin{equation*}
u_t + \mathcal{N}[u;\lambda] = 0,
\end{equation*}
where $u(t,x)$ denotes the latent (hidden) solution and $\mathcal{N}[\cdot;\lambda]$ is a nonlinear operator parametrized by $\lambda$. This setup encapsulates a wide range of problems in mathematical physics including conservation laws, diffusion processes, advection-diffusion-reaction systems, and kinetic equations. As a motivating example, the one dimensional Burgers' equation \cite{basdevant1986spectral} corresponds to the case where $\mathcal{N}[u;\lambda] = \lambda_1 u u_x - \lambda_2 u_{xx}$ and $\lambda = (\lambda_1, \lambda_2)$. Here, the subscripts denote partial differentiation in either time or space. Given noisy measurements of the system, we are interested in the solution of two distinct problems. The first problem is that of predictive inference, filtering and smoothing, or data driven solutions of partial differential equations \cite{raissi2017numerical, raissi2017inferring} which states: given fixed model parameters $\lambda$ what can be said about the unknown hidden state $u(t,x)$ of the system? The second problem is that of learning, system identification, or data-driven discovery of partial differential equations \cite{raissi2017hidden,raissi2017machine, Rudye1602614} stating: what are the parameters $\lambda$ that best describe the observed data?\\

In this first part of our two-part treatise, we focus on computing data-driven solutions to partial differential equations of the general form
\begin{eqnarray}\label{eq:PDE}
&&u_t + \mathcal{N}[u] = 0,\ x \in \Omega, \ t\in[0,T],
\end{eqnarray}
where $u(t,x)$ denotes the latent (hidden) solution, $\mathcal{N}[\cdot]$ is a nonlinear differential operator, and $\Omega$ is a subset of $\mathbb{R}^D$. In what follows, we put forth two distinct classes of algorithms, namely continuous and discrete time models, and highlight their properties and performance through the lens of different benchmark problems. All code and data-sets accompanying this manuscript are available at \url{https://github.com/maziarraissi/PINNs}.

\section{Continuous Time Models}
We define $f(t,x)$ to be given by the left-hand-side of equation \eqref{eq:PDE}; i.e.,
\begin{equation}
f := u_t + \mathcal{N}[u],\label{eq:PDE_RHS}
\end{equation}
and proceed by approximating $u(t,x)$ by a deep neural network. This assumption along with equation \eqref{eq:PDE_RHS} result in a \emph{physics informed neural network} $f(t,x)$. This network can be derived by applying the chain rule for differentiating compositions of functions using automatic differentiation \cite{baydin2015automatic}.

\subsection{Example (Burgers' Equation)}
As an example, let us consider the Burgers' equation. This equation arises in various areas of applied mathematics, including fluid mechanics, nonlinear acoustics, gas dynamics, and traffic flow \cite{basdevant1986spectral}. It is a fundamental partial differential equation and can be derived from the Navier-Stokes equations for the velocity field by dropping the pressure gradient term. For small values of the viscosity parameters, Burgers' equation can lead to shock formation that is notoriously hard to resolve by classical numerical methods. In one space dimension, the Burger's equation along with Dirichlet boundary conditions reads as
\begin{eqnarray}\label{eq:Burgers}
&& u_t + u u_x - (0.01/\pi) u_{xx} = 0,\ \ \ x \in [-1,1],\ \ \ t \in [0,1],\\
&& u(0,x) = -\sin(\pi x),\nonumber\\
&& u(t,-1) = u(t,1) = 0.\nonumber
\end{eqnarray}
Let us define $f(t,x)$ to be given by
\[
f := u_t + u u_x - (0.01/\pi) u_{xx},
\]
and proceed by approximating $u(t,x)$ by a deep neural network. To highlight the simplicity in implementing this idea we have included a  Python code snippet using Tensorflow \cite{abadi2016tensorflow}; currently one of the most popular and well documented open source libraries for machine learning computations. To this end, $u(t,x)$ can be simply defined as
\begin{lstlisting}[language=Python]
def u(t, x):
    u = neural_net(tf.concat([t,x],1), weights, biases)
    return u
\end{lstlisting}
Correspondingly, the \emph{physics informed neural network} $f(t,x)$ takes the form
\begin{lstlisting}[language=Python]
def f(t, x):
    u = u(t, x)
    u_t = tf.gradients(u, t)[0]
    u_x = tf.gradients(u, x)[0]
    u_xx = tf.gradients(u_x, x)[0]
    f = u_t + u*u_x - (0.01/tf.pi)*u_xx
    return f
\end{lstlisting}
The shared parameters between the neural networks $u(t,x)$ and $f(t,x)$ can be learned by minimizing the mean squared error loss
\begin{equation}\label{eq:MSE_Burgers_CT_inference}
MSE = MSE_u + MSE_f,
\end{equation}
where
\[
MSE_u = \frac{1}{N_u}\sum_{i=1}^{N_u} |u(t^i_u,x_u^i) - u^i|^2,
\]
and
\[
MSE_f = \frac{1}{N_f}\sum_{i=1}^{N_f}|f(t_f^i,x_f^i)|^2.
\]
Here, $\{t_u^i, x_u^i, u^i\}_{i=1}^{N_u}$ denote the initial and boundary training data on $u(t,x)$ and $\{t_f^i, x_f^i\}_{i=1}^{N_f}$ specify the collocations points for $f(t,x)$. The loss $MSE_u$ corresponds to the initial and boundary data while $MSE_f$ enforces the structure imposed by equation \eqref{eq:Burgers} at a finite set of collocation points.\\

In all benchmarks considered in this work, the total number of training data $N_u$ is relatively small (a few hundred up to a few thousand points), and we chose to optimize all loss functions using L-BFGS; a quasi-Newton, full-batch gradient-based optimization algorithm \cite{liu1989limited}. For larger data-sets a more computationally efficient mini-batch setting can be readily employed using stochastic gradient descent and its modern variants \cite{goodfellow2016deep,kingma2014adam}. Despite the fact that there is no theoretical guarantee that this procedure converges to a global minimum, our empirical evidence indicates that, if the given partial differential equation is well-posed and its solution is unique, our method is capable of achieving good prediction accuracy given a sufficiently expressive neural network architecture and a sufficient number of collocation points $N_f$. This general observation deeply relates to the resulting optimization landscape induced by the mean square error loss of equation \ref{eq:MSE_Burgers_CT_inference}, and defines an open question for research that is in sync with recent theoretical developments in deep learning \cite{choromanska2015loss,shwartz2017opening}.
Here, we will test the robustness of the proposed methodology using a series of systematic sensitivity studies that accompany the numerical results presented in the following.\\

\begin{figure}[!t]
\includegraphics[width = 1.0\textwidth]{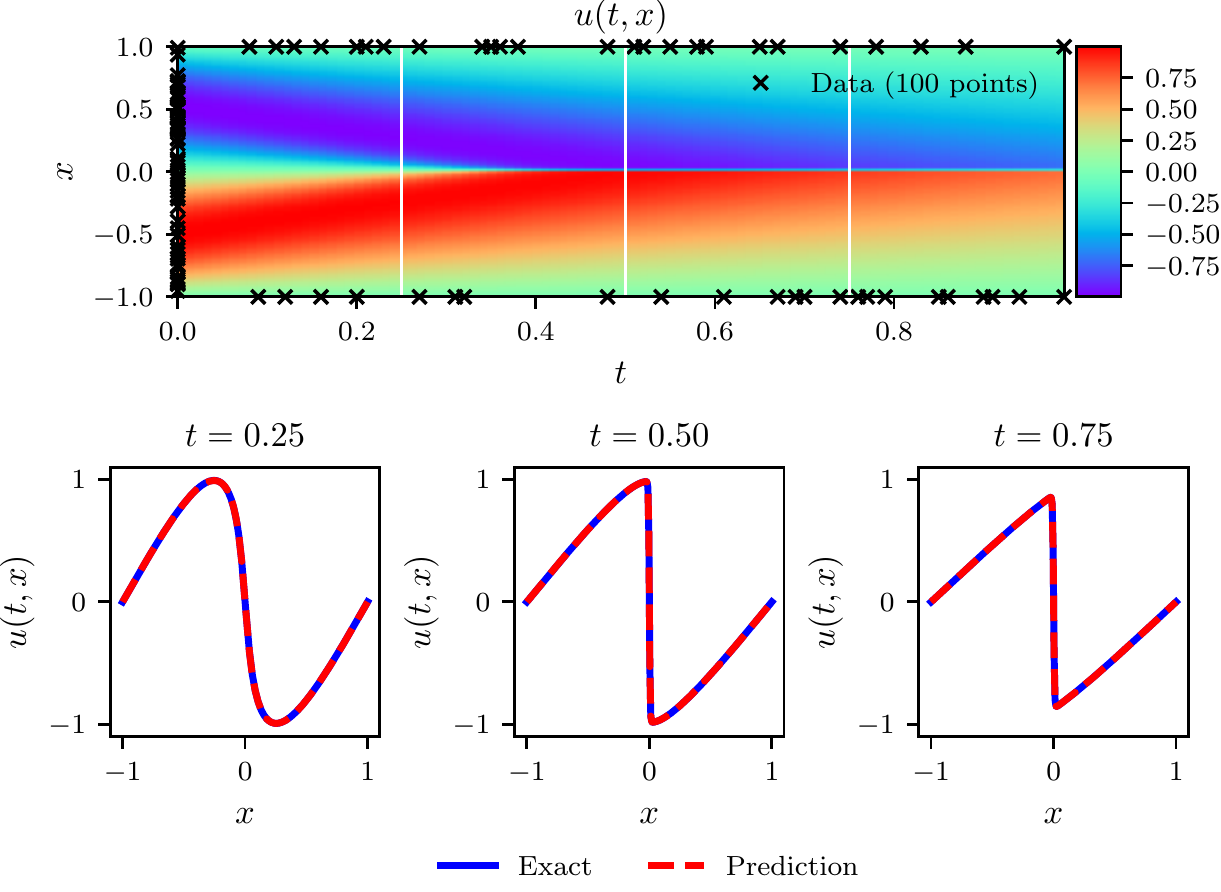}
\caption{{\em Burgers' equation:} {\it Top:} Predicted solution $u(t,x)$ along with the initial and boundary training data. In addition we are using 10,000 collocation points generated using a Latin Hypercube Sampling strategy. {\it Bottom:} Comparison of the predicted and exact solutions corresponding to the three temporal snapshots depicted by the white vertical lines in the top panel. The relative $\mathcal{L}_{2}$ error for this case is $6.7 \cdot 10^{-4}$. Model training took approximately 60 seconds on a single NVIDIA Titan X GPU card.}
\label{fig:Burgers_CT_inference}
\end{figure}

Figure \ref{fig:Burgers_CT_inference} summarizes our results for the data-driven solution of the Burgers equation. Specifically, given a set of $N_u = 100$ randomly distributed initial and boundary data, we learn the latent solution $u(t,x)$ by training all $3021$ parameters of a 9-layer deep neural network using the mean squared error loss of \eqref{eq:MSE_Burgers_CT_inference}. Each hidden layer contained $20$ neurons and a hyperbolic tangent activation function. In general, the neural network should be given sufficient approximation capacity in order to accommodate the anticipated complexity of $u(t,x)$. However, in this example, our choice aims to highlight the robustness of the proposed method with respect to the well known issue of over-fitting. Specifically, the term in $MSE_f$ in equation \eqref{eq:MSE_Burgers_CT_inference} acts as a regularization mechanism that penalizes solutions that do not satisfy equation \eqref{eq:Burgers}. Therefore, a key property of {\em physics informed neural networks} is that they can be effectively trained using small data sets; a setting often encountered in the study of physical systems for which the cost of data acquisition may be prohibitive.\\

The top panel of Figure \ref{fig:Burgers_CT_inference} shows the predicted spatio-temporal solution $u(t,x)$, along with the locations of the initial and boundary training data. We must underline that, unlike any classical numerical method for solving partial differential equations, this prediction is obtained without any sort of discretization of the spatio-temporal domain. The exact solution for this problem is analytically available \cite{basdevant1986spectral}, and the resulting prediction error is measured at $6.7 \cdot 10^{-4}$ in the relative $\mathcal{L}_2$-norm. Note that this error is about two orders of magnitude lower than the one reported in our previous work on data-driven solution of partial differential equation using Gaussian processes \cite{raissi2017numerical}. A more detailed assessment of the predicted solution is presented in the bottom panel of figure \ref{fig:Burgers_CT_inference}. In particular, we present a comparison between the exact and the predicted solutions at different time instants $t=0.25,0.50,0.75$. Using only a handful of initial and boundary data, the {\em physics informed neural network} can accurately capture the intricate nonlinear behavior of the Burgers' equation that leads to the development of a sharp internal layer around $t = 0.4$. The latter is notoriously hard to accurately resolve with classical numerical methods and requires a laborious spatio-temporal discretization of equation \eqref{eq:Burgers}.\\

To further analyze the performance of our method, we have performed the following systematic studies to quantify its predictive accuracy for different number of training and collocation points, as well as for different neural network architectures. In table \ref{tab:Burgers_CT_inference_1} we report the resulting relative $\mathcal{L}_{2}$ error for different number of initial and boundary training data $N_u$ and different number of collocation points $N_f$, while keeping the 9-layer network architecture fixed. The general trend shows increased prediction accuracy as the total number of training data $N_u$ is increased, given a sufficient number of collocation points $N_f$. This observation highlights a key strength of {\em physics informed neural networks}: by encoding the structure of the underlying physical law through the collocation points $N_f$, one can obtain a more accurate and data-efficient learning algorithm.\footnote{Note that the case $N_f = 0$ corresponds to a standard neural network model, i.e., a neural network that does not take into account the underlying governing equation.} Finally, table \ref{tab:Burgers_CT_inference_2} shows the resulting relative $\mathcal{L}_{2}$ for different number of hidden layers, and different number of neurons per layer, while the total number of training and collocation points is kept fixed to $N_u = 100$ and $N_f=10,000$, respectively. As expected, we observe that as the number of layers and neurons is increased (hence the capacity of the neural network to approximate more complex functions), the predictive accuracy is increased.

\begin{table}[!t]
\centering
\begin{tabular}{|l||cccccc|} 
\hline
\diagbox{$N_u$}{$N_f$} & 2000 & 4000 & 6000 & 7000 & 8000 & 10000 \\ \hline\hline
20 & 2.9e-01 & 4.4e-01 & 8.9e-01 & 1.2e+00 & 9.9e-02 & 4.2e-02 \\ 
40 & 6.5e-02 & 1.1e-02 & 5.0e-01 & 9.6e-03 & 4.6e-01 & 7.5e-02 \\ 
60 & 3.6e-01 & 1.2e-02 & 1.7e-01 & 5.9e-03 & 1.9e-03 & 8.2e-03 \\ 
80 & 5.5e-03 & 1.0e-03 & 3.2e-03 & 7.8e-03 & 4.9e-02 & 4.5e-03 \\ 
100 & 6.6e-02 & 2.7e-01 & 7.2e-03 & 6.8e-04 & 2.2e-03 & 6.7e-04 \\ 
200 & 1.5e-01 & 2.3e-03 & 8.2e-04 & 8.9e-04 & 6.1e-04 & 4.9e-04 \\ \hline
\end{tabular}
\caption{{\em Burgers' equation:} Relative $\mathcal{L}_{2}$ error between the predicted and the exact solution $u(t,x)$ for different number of initial and boundary training data $N_u$, and different number of collocation points $N_f$. Here, the network architecture is fixed to 9 layers with 20 neurons per hidden layer.} \label{tab:Burgers_CT_inference_1}
\end{table}

\begin{table}[!t]
\centering
\begin{tabular}{|c||ccc|} 
\hline
\diagbox{Layers}{Neurons} & 10 & 20 & 40  \\ \hline\hline
2 & 7.4e-02 & 5.3e-02 & 1.0e-01 \\ 
4 & 3.0e-03 & 9.4e-04 & 6.4e-04 \\ 
6 & 9.6e-03 & 1.3e-03 & 6.1e-04 \\ 
8 & 2.5e-03 & 9.6e-04 & 5.6e-04 \\ \hline
\end{tabular}
\caption{{\em Burgers' equation:} Relative $\mathcal{L}_{2}$ error between the predicted and the exact solution $u(t,x)$ for different number of hidden layers and different number of neurons per layer. Here, the total number of training and collocation points is fixed to $N_u = 100$ and $N_f=10,000$, respectively.} \label{tab:Burgers_CT_inference_2}
\end{table}

\subsection{Example (Shr\"{o}dinger Equation)} \label{sec:schrodinger_CT}
This example aims to highlight the ability of our method to handle periodic boundary conditions, complex-valued solutions, as well as different types of nonlinearities in the governing partial differential equations. The one-dimensional nonlinear Schr\"{o}dinger equation is a classical field equation that is used to study quantum mechanical systems, including nonlinear wave propagation in optical fibers and/or waveguides, Bose-Einstein condensates, and plasma waves. In optics, the nonlinear term arises from the intensity dependent index of refraction of a given material. Similarly, the nonlinear term for Bose-Einstein condensates is a result of the mean-field interactions of an interacting, N-body system. The nonlinear Schr\"{o}dinger equation along with periodic boundary conditions is given by
\begin{eqnarray}\label{eq:Schrodinger}
&& i h_t + 0.5 h_{xx} + |h|^2 h = 0,\ \ \ x \in [-5, 5],\ \ \ t \in [0, \pi/2],\\
&& h(0,x) = 2\ \text{sech}(x),\nonumber\\
&& h(t,-5) = h(t, 5),\nonumber\\
&& h_x(t,-5) = h_x(t, 5),\nonumber
\end{eqnarray}
where $h(t,x)$ is the complex-valued solution. Let us define $f(t,x)$ to be given by
\[
f := i h_t + 0.5 h_{xx} + |h|^2 h,
\]
and proceed by placing a complex-valued neural network prior on $h(t,x)$. In fact, if $u$ denotes the real part of $h$ and $v$ is the imaginary part, we are placing a multi-out neural network prior on $h(t,x) = \begin{bmatrix}
u(t,x) & v(t,x)
\end{bmatrix}$. This will result in the complex-valued (multi-output) \emph{physic informed neural network} $f(t,x)$. The shared parameters of the neural networks $h(t,x)$ and $f(t,x)$ can be learned by minimizing the mean squared error loss
\begin{equation}\label{eq:MSE_Schrodinger}
MSE = MSE_0 + MSE_b + MSE_f,
\end{equation}
where
\[
MSE_0 = \frac{1}{N_0}\sum_{i=1}^{N_0} |h(0,x_0^i) - h^i_0|^2,
\]
\[
MSE_b = \frac{1}{N_b}\sum_{i=1}^{N_b} \left(|h^i(t^i_b,-5) - h^i(t^i_b,5)|^2 + |h^i_x(t^i_b,-5) - h^i_x(t^i_b,5)|^2\right),
\]
and
\[
MSE_f = \frac{1}{N_f}\sum_{i=1}^{N_f}|f(t_f^i,x_f^i)|^2.
\]
Here, $\{x_0^i, h^i_0\}_{i=1}^{N_0}$ denotes the initial data, $\{t^i_b\}_{i=1}^{N_b}$ corresponds to the collocation points on the boundary, and $\{t_f^i,x_f^i\}_{i=1}^{N_f}$ represents the collocation points on $f(t,x)$. Consequently, $MSE_0$ corresponds to the loss on the initial data, $MSE_b$ enforces the periodic boundary conditions, and $MSE_f$ penalizes the Schr\"{o}dinger equation not being satisfied on the collocation points.\\

In order to assess the accuracy of our method, we have simulated equation \eqref{eq:Schrodinger} using conventional spectral methods to create a high-resolution data set. Specifically, starting from an initial state $h(0,x) = 2\ \text{sech}(x)$ and assuming periodic boundary conditions $h(t,-5) = h(t,5)$ and $h_x(t,-5) = h_x(t,5)$, we have integrated equation \eqref{eq:Schrodinger} up to a final time $t=\pi/2$ using the Chebfun package \cite{driscoll2014chebfun} with a spectral Fourier discretization with 256 modes and a fourth-order explicit Runge-Kutta temporal integrator with time-step $\Delta{t} = \pi/2 \cdot 10^{-6}$. Under our data-driven setting, all we observe are measurements $\{x_0^i, h^i_0\}_{i=1}^{N_0}$ of the latent function $h(t,x)$ at time $t=0$. In particular, the training set consists of a total of $N_0 = 50$ data points on $h(0,x)$ randomly parsed from the full high-resolution data-set, as well as $N_b = 50$ randomly sampled collocation points $\{t^i_b\}_{i=1}^{N_b}$ for enforcing the periodic boundaries. Moreover, we have assumed $N_f=20,000$ randomly sampled collocation points used to enforce equation \eqref{eq:Schrodinger} inside the solution domain. All randomly sampled point locations were generated using a space filling Latin Hypercube Sampling strategy \cite{stein1987large}.\\

\begin{figure}[!t]
\includegraphics[width = 1.0\textwidth]{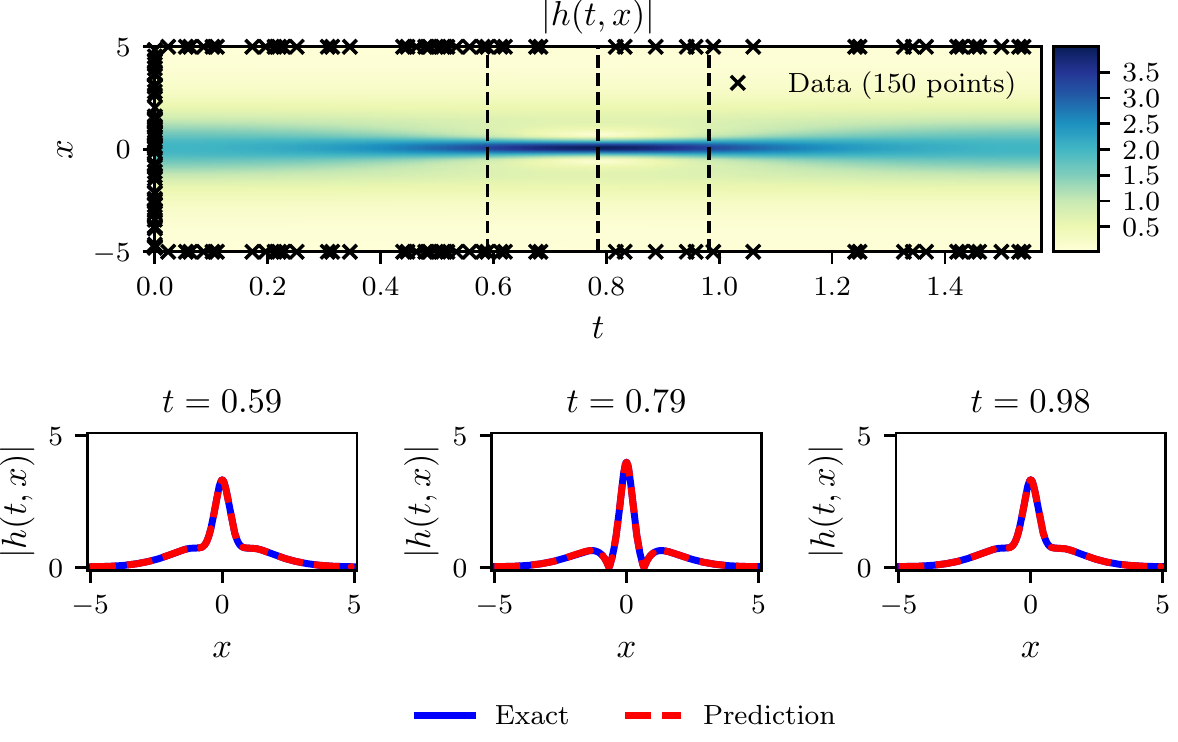}
\caption{{\em Shr\"{o}dinger equation:} {\it Top:} Predicted solution $|h(t,x)|$ along with the initial and boundary training data. In addition we are using 20,000 collocation points generated using a Latin Hypercube Sampling strategy. {\it Bottom:} Comparison of the predicted and exact solutions corresponding to  the three temporal snapshots depicted by the dashed vertical lines in the top panel. The relative $\mathcal{L}_{2}$ error for this case is $1.97 \cdot 10^{-3}$.}
\label{fig:NLS}
\end{figure}

Here our goal is to infer the entire spatio-temporal solution $h(t,x)$ of the Schr\"{o}dinger equation (\ref{eq:Schrodinger}). We chose to jointly represent the latent function $h(t,x) = [u(t,x)\ v(t,x)]$ using a 5-layer deep neural network with $100$ neurons per layer and a hyperbolic tangent activation function. Figure \ref{fig:NLS} summarizes the results of our experiment. Specifically, the top panel of figure \ref{fig:NLS} shows the magnitude of the predicted spatio-temporal solution $|h(t,x)|=\sqrt{u^2(t,x) + v^2(t,x)}$, along with the locations of the initial and boundary training data. The resulting prediction error is validated against the test data for this problem, and is measured at $1.97 \cdot 10^{-3}$ in the relative $\mathcal{L}_2$-norm. A more detailed assessment of the predicted solution is presented in the bottom panel of Figure~\ref{fig:NLS}. In particular, we present a comparison between the exact and the predicted solutions at different time instants $t=0.59,0.79,0.98$. Using only a handful of initial data, the {\em physics informed neural network} can accurately capture the intricate nonlinear behavior of the Schr\"{o}dinger equation.\\

One potential limitation of the continuous time neural network models considered so far, stems from the need to use a large number of collocation points $N_f$ in order to enforce physics informed constraints in the entire spatio-temporal domain. Although this poses no significant issues for problems in one or two spatial dimensions, it may introduce a severe bottleneck in higher dimensional problems, as the total number of collocation points needed to globally enforce a physics informed constrain (i.e., in our case a partial differential equation) will increase exponentially. In the next section, we put forth a different approach that circumvents the need for collocation points by introducing a more structured neural network representation leveraging the classical Runge-Kutta time-stepping schemes \cite{iserles2009first}.

\section{Discrete Time Models}\label{sec:DT_models}
Let us apply the general form of Runge-Kutta methods with $q$ stages \cite{iserles2009first} to equation (\ref{eq:PDE}) and obtain
\begin{equation}\label{eq:RungeKutta}
\arraycolsep=1.0pt\def\arraystretch{1.5}
\begin{array}{ll}
u^{n+c_i} = u^n - \Delta t \sum_{j=1}^q a_{ij} \mathcal{N}[u^{n+c_j}], \ \ i=1,\ldots,q,\\
u^{n+1} = u^{n} - \Delta t \sum_{j=1}^q b_j \mathcal{N}[u^{n+c_j}].
\end{array}
\end{equation}
Here, $u^{n+c_j}(x) = u(t^n + c_j \Delta t, x)$ for $j=1, \ldots, q$. This general form encapsulates both implicit and explicit time-stepping schemes, depending on the choice of the parameters $\{a_{ij},b_j,c_j\}$. Equations (\ref{eq:RungeKutta}) can be equivalently expressed as
\begin{equation}
\arraycolsep=1.0pt\def\arraystretch{1.5}
\begin{array}{ll}
u^{n} = u^n_i, \ \ i=1,\ldots,q,\\
u^n = u^n_{q+1},
\end{array}
\end{equation}
where
\begin{equation}\label{eq:RungeKutta_inference_rearranged}
\arraycolsep=1.0pt\def\arraystretch{1.5}
\begin{array}{ll}
u^n_i := u^{n+c_i} + \Delta t \sum_{j=1}^q a_{ij} \mathcal{N}[u^{n+c_j}], \ \ i=1,\ldots,q,\\
u^n_{q+1} := u^{n+1} + \Delta t \sum_{j=1}^q b_j \mathcal{N}[u^{n+c_j}].
\end{array}
\end{equation}
We proceed by placing a multi-output neural network prior on
\begin{equation}\label{eq:RungeKutta_PU_prior_inference}
\begin{bmatrix}
u^{n+c_1}(x), \ldots, u^{n+c_q}(x), u^{n+1}(x)
\end{bmatrix}.
\end{equation}
This prior assumption along with equations (\ref{eq:RungeKutta_inference_rearranged}) result in a \emph{physics informed neural network} that takes $x$ as an input and outputs
\begin{equation}\label{eq:RungeKutta_PI_prior_inference}
\begin{bmatrix}
u^n_1(x), \ldots, u^n_q(x), u^n_{q+1}(x)
\end{bmatrix}.
\end{equation}

\subsection{Example (Burgers' Equation)}
To highlight the key features of the discrete time representation we revisit the problem of data-driven solution of the Burgers' equation. For this case, the nonlinear operator in equation \eqref{eq:RungeKutta_inference_rearranged} is given by
\[
\mathcal{N}[u^{n+c_j}] = u^{n+c_j} u^{n+c_j}_x - (0.01/\pi)u^{n+c_j}_{xx},
\]
and the shared parameters of the neural networks \eqref{eq:RungeKutta_PU_prior_inference} and \eqref{eq:RungeKutta_PI_prior_inference} can be learned by minimizing the sum of squared errors
\begin{equation}\label{eq:SSE_Burgers_DT_inference}
SSE = SSE_n + SSE_{b},
\end{equation}
where
\[
SSE_n = \sum_{j=1}^{q+1} \sum_{i=1}^{N_n} |u^n_j(x^{n,i}) - u^{n,i}|^2,
\]
and
\[
SSE_b = \sum_{i=1}^q \left(|u^{n+c_i}(-1)|^2 + |u^{n+c_i}(1)|^2\right) + |u^{n+1}(-1)|^2 + |u^{n+1}(1)|^2.
\]
Here, $\{x^{n,i}, u^{n,i}\}_{i=1}^{N_n}$ corresponds to the data at time $t^n$. The Runge-Kutta scheme now allows us to infer the latent solution $u(t,x)$ in a sequential fashion. Starting from initial data $\{x^{n,i}, u^{n,i}\}_{i=1}^{N_n}$ at time $t^n$ and data at the domain boundaries $x = -1$ and $x = 1$, we can use the aforementioned loss function \eqref{eq:SSE_Burgers_DT_inference} to train the networks of \eqref{eq:RungeKutta_PU_prior_inference}, \eqref{eq:RungeKutta_PI_prior_inference}, and predict the solution at time $t^{n+1}$. A Runge-Kutta time-stepping scheme would then use this prediction as initial data for the next step and proceed to train again and predict $u(t^{n+2},x)$, $u(t^{n+3},x)$, etc., one step at a time.\\

In classical numerical analysis, these steps are usually confined to be small due to stability constraints for explicit schemes or computational complexity constrains for implicit formulations \cite{iserles2009first}. These constraints become more severe as the total number of Runge-Kutta stages $q$ is increased, and, for most problems of practical interest, one needs to take thousands to millions of such steps until the solution is resolved up to a desired final time. In sharp contrast to classical methods, here we can employ implicit Runge-Kutta schemes with an arbitrarily large number of stages at effectively no extra cost.\footnote{To be precise, it is only the number of parameters in the last layer of the neural network that increases linearly with the total number of stages.} This enables us to take very large time steps while retaining stability and high predictive accuracy, therefore allowing us to resolve the entire spatio-temporal solution in a single step.\\

\begin{figure}[!t]
\includegraphics[width = 1.0\textwidth]{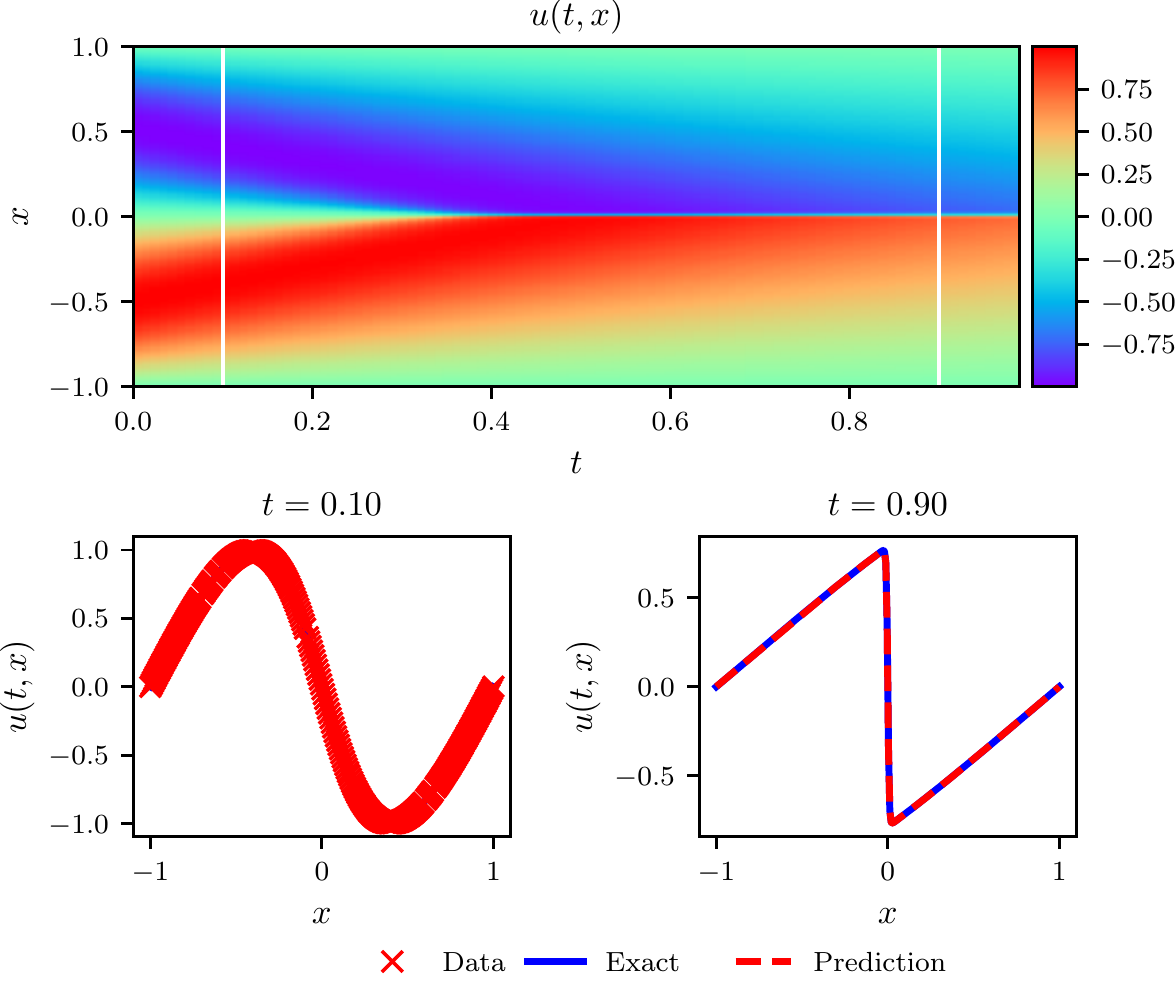}
\caption{{\em Burgers equation:} {\it Top:} Solution $u(t,x)$ along with the location of the initial training snapshot at $t=0.1$ and the final prediction snapshot at $t=0.9$. {\it Bottom:} Initial training data and final prediction at the snapshots depicted by the white vertical lines in the top panel. The relative $\mathcal{L}_{2}$ error for this case is $8.2 \cdot 10^{-4}$.}
\label{fig:Burgers_DT_inference}
\end{figure}

The result of applying this process to the Burgers' equation is presented in figure \ref{fig:Burgers_DT_inference}. For illustration purposes, we start with a set of $N_n=250$ initial data at $t = 0.1$, and employ a {\em physics informed neural network} induced by an implicit Runge-Kutta scheme with 500 stages to predict the solution at time $t=0.9$ in a single step. The theoretical error estimates for this scheme predict a temporal error accumulation of $\mathcal{O}(\Delta{t}^{2q})$ \cite{iserles2009first}, which in our case translates into an error way below machine precision, i.e., $\Delta{t}^{2q} = 0.8^{1000} \approx 10^{-97}$. To our knowledge, this is the first time that an implicit Runge-Kutta scheme of that high-order has ever been used. Remarkably, starting from smooth initial data at $t=0.1$ we can predict the nearly discontinuous solution at $t=0.9$ in a single time-step with a relative $\mathcal{L}_{2}$ error of $8.2 \cdot 10^{-4}$. This error is two orders of magnitude lower that the one reported in \cite{raissi2017numerical}, and it is entirely attributed to the neural network's capacity to approximate $u(t,x)$, as well as to the degree that the sum of squared errors loss allows interpolation of the training data. The network architecture used here consists of 4 layers with 50 neurons in each hidden layer.\\

A detailed systematic study to quantify the effect of different network architectures is presented in table \ref{tab:Burgers_DT_inference_2}. By keeping the number of Runge-Kutta stages fixed to $q = 500$ and the time-step size to $\Delta{t}=0.8$, we have varied the number of hidden layers and the number of neurons per layer, and monitored the resulting relative $\mathcal{L}_{2}$ error for the predicted solution at time $t=0.9$. Evidently, as the neural network capacity is increased the predictive accuracy is enhanced.\\

\begin{table}[!t]
\centering
\begin{tabular}{|c||ccc|} 
\hline
\diagbox{Layers}{Neurons} & 10 & 25 & 50  \\ \hline\hline
1 & 4.1e-02 & 4.1e-02 & 1.5e-01 \\
2 & 2.7e-03 & 5.0e-03 & 2.4e-03 \\
3 & 3.6e-03 & 1.9e-03 & 9.5e-04 \\ \hline
\end{tabular}
\caption{{\em Burgers' equation:} Relative final prediction error measure in the $\mathcal{L}_{2}$ norm for different number of hidden layers and neurons in each layer. Here, the number of Runge-Kutta stages is fixed to 500 and the time-step size to $\Delta{t}=0.8$.} \label{tab:Burgers_DT_inference_2}
\end{table}

The key parameters controlling the performance of our discrete time algorithm are the total number of Runge-Kutta stages $q$ and the time-step size $\Delta{t}$. In table \ref{tab:Burgers_DT_inference_1} we summarize the results of an extensive systematic study where we fix the network architecture to 4 hidden layers with 50 neurons per layer, and vary the number of Runge-Kutta stages $q$ and the time-step size $\Delta{t}$. Specifically, we see how cases with low numbers of stages fail to yield accurate results when the time-step size is large. For instance, the case $q=1$ corresponding to the classical trapezoidal rule, and the case $q=2$ corresponding to the $4^{\text{th}}$-order Gauss-Legendre method, cannot retain their predictive accuracy for time-steps larger than 0.2, thus mandating a solution strategy with multiple time-steps of small size. On the other hand, the ability to push the number of Runge-Kutta stages to 32 and even higher allows us to take very large time steps, and effectively resolve the solution in a single step without sacrificing the accuracy of our predictions. Moreover, numerical stability is not sacrificed either as implicit Runge-Kutta is the only family of time-stepping schemes that remain A-stable regardless of their order, thus making them ideal for stiff problems \cite{iserles2009first}. These properties are unprecedented for an algorithm of such implementation simplicity, and illustrate one of the key highlights of our discrete time approach.

\begin{table}[!t]
\centering
\begin{tabular}{|l||cccc|} 
\hline
\diagbox{$q$}{$\Delta{t}$} & 0.2 & 0.4 & 0.6 & 0.8 \\ \hline\hline
1 & 3.5e-02 & 1.1e-01 & 2.3e-01 & 3.8e-01 \\
2 & 5.4e-03 & 5.1e-02 & 9.3e-02 & 2.2e-01 \\
4 & 1.2e-03 & 1.5e-02 & 3.6e-02 & 5.4e-02 \\
8 & 6.7e-04 & 1.8e-03 & 8.7e-03 & 5.8e-02 \\
16 & 5.1e-04 & 7.6e-02 & 8.4e-04 & 1.1e-03 \\
32 & 7.4e-04 & 5.2e-04 & 4.2e-04 & 7.0e-04 \\
64 & 4.5e-04 & 4.8e-04 & 1.2e-03 & 7.8e-04 \\
100 & 5.1e-04 & 5.7e-04 & 1.8e-02 & 1.2e-03 \\
500 & 4.1e-04 & 3.8e-04 & 4.2e-04 & 8.2e-04 \\ \hline
\end{tabular}
\caption{{\em Burgers' equation:} Relative final prediction error measured in the $\mathcal{L}_{2}$ norm for different number of Runge-Kutta stages $q$ and time-step sizes $\Delta{t}$. Here, the network architecture is fixed to 4 hidden layers with 50 neurons in each layer.} \label{tab:Burgers_DT_inference_1}
\end{table}

\subsubsection{Example (Allen-Cahn Equation)}
This example aims to highlight the ability of the proposed discrete time models to handle different types of nonlinearity in the governing partial differential equation. To this end, let us consider the Allen-Cahn equation along with periodic boundary conditions
\begin{eqnarray} \label{eq:Allen-Cahn}
&&u_t - 0.0001 u_{xx} + 5 u^3 - 5 u = 0, \ \ \ x \in [-1,1], \ \ \ t \in [0,1],\\
&&u(0, x) = x^2 \cos(\pi x),\nonumber\\
&&u(t,-1) = u(t,1),\nonumber\\
&&u_x(t,-1) = u_x(t,1).\nonumber
\end{eqnarray}
The Allen-Cahn equation is a well-known equation from the area of reaction-diffusion systems. It describes the process of phase separation in multi-component alloy systems, including order-disorder transitions. For the Allen-Cahn equation, the nonlinear operator in equation \eqref{eq:RungeKutta_inference_rearranged} is given by
\[
\mathcal{N}[u^{n+c_j}] = -0.0001 u^{n+c_j}_{xx} + 5 \left(u^{n+c_j}\right)^3 - 5 u^{n+c_j},
\]
and the shared parameters of the neural networks \eqref{eq:RungeKutta_PU_prior_inference} and \eqref{eq:RungeKutta_PI_prior_inference} can be learned by minimizing the sum of squared errors
\begin{equation}\label{eq:SSE_Allen-Cahn}
SSE = SSE_n + SSE_b,
\end{equation}
where
\[
SSE_n = \sum_{j=1}^{q+1} \sum_{i=1}^{N_n} |u^n_j(x^{n,i}) - u^{n,i}|^2,
\]
and
\begin{eqnarray*}
SSE_b &=& \sum_{i=1}^q |u^{n+c_i}(-1) - u^{n+c_i}(1)|^2 + |u^{n+1}(-1) - u^{n+1}(1)|^2 \\
      &+& \sum_{i=1}^q |u_x^{n+c_i}(-1) - u_x^{n+c_i}(1)|^2 + |u_x^{n+1}(-1) - u_x^{n+1}(1)|^2.
\end{eqnarray*}
Here, $\{x^{n,i}, u^{n,i}\}_{i=1}^{N_n}$ corresponds to the data at time $t^n$. We have generated a training and test data-set set by simulating the Allen-Cahn equation \eqref{eq:Allen-Cahn} using conventional spectral methods. Specifically, starting from an initial condition $u(0,x) = x^2 \cos(\pi x)$ and assuming periodic boundary conditions $u(t,-1) = u(t,1)$ and $u_x(t,-1) = u_x(t,1)$, we have integrated equation \eqref{eq:Allen-Cahn} up to a final time $t=1.0$ using the Chebfun package \cite{driscoll2014chebfun} with a spectral Fourier discretization with 512 modes and a fourth-order explicit Runge-Kutta temporal integrator with time-step $\Delta{t} = 10^{-5}$.\\

\begin{figure}[!t]
\includegraphics[width = 1.0\textwidth]{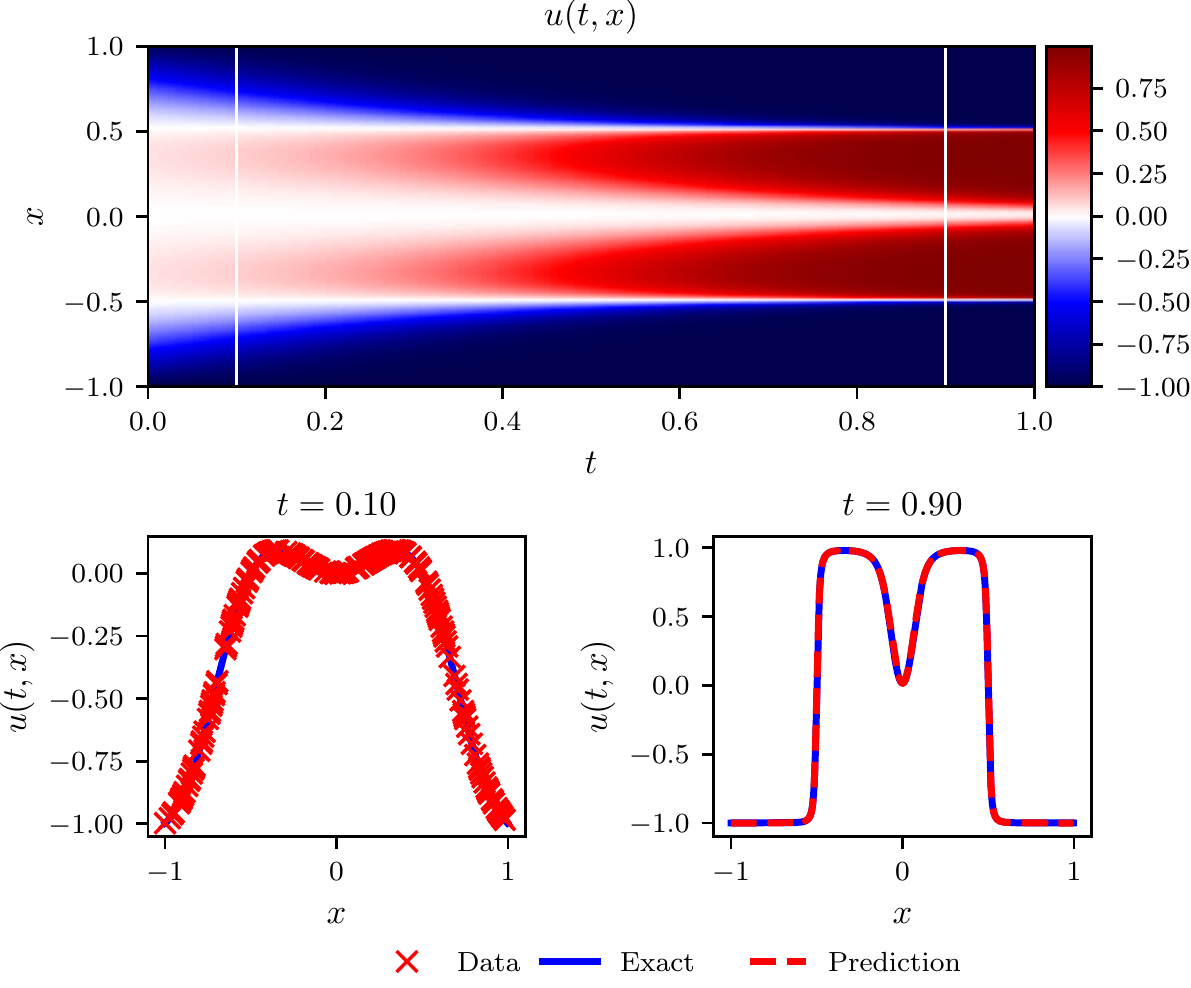}
\caption{{\em Allen-Cahn equation:} 
{\it Top:} Solution $u(t,x)$ along with the location of the initial training snapshot at $t=0.1$ and the final prediction snapshot at $t=0.9$. {\it Bottom:} Initial training data and final prediction at the snapshots depicted by the white vertical lines in the top panel. The relative $\mathcal{L}_{2}$ error for this case is $6.99\cdot 10^{-3}$.}
\label{fig:AC_DT_inference}
\end{figure}

In this example, we assume $N_n = 200$ initial data points that are randomly sub-sampled from the exact solution at time $t=0.1$, and our goal is to predict the solution at time $t=0.9$ using a single time-step with size $\Delta{t}=0.8$. To this end, we employ a discrete time {\em physics informed neural network} with 4 hidden layers and 200 neurons per layer, while the output layer predicts 101 quantities of interest corresponding to the $q=100$ Runge-Kutta stages $u^{n+c_i}(x)$, $i=1,\dots,q$, and the solution at final time $u^{n+1}(x)$. Figure \ref{fig:AC_DT_inference} summarizes our predictions after the network has been trained using the loss function of equation \eqref{eq:SSE_Allen-Cahn}. Evidently, despite the complex dynamics leading to a solution with two sharp internal layers, we are able to obtain an accurate prediction of the solution at $t=0.9$ using only a small number of scattered measurements at $t=0.1$.

\section{Summary and Discussion} 

We have introduced {\em physics informed neural networks}, a new class of universal function approximators that is capable of encoding any underlying physical laws that govern a given data-set, and can be described by partial differential equations. In this work, we design data-driven algorithms for inferring solutions to general nonlinear partial differential equations, and constructing computationally efficient physics-informed surrogate models. The resulting methods showcase a series of promising results for a diverse collection of problems in computational science, and open the path for endowing deep learning with the powerful capacity of mathematical physics to model the world around us. As deep learning technology is continuing to grow rapidly both in terms of methodological and algorithmic developments, we believe that this is a timely contribution that can benefit practitioners across a wide range of scientific domains. Specific applications that can readily enjoy these benefits include, but are not limited to, data-driven forecasting of physical processes, model predictive control, multi-physics/multi-scale modeling and simulation.\\

We must note however that the proposed methods should not be viewed as replacements of classical numerical methods for solving partial differential equations (e.g., finite elements, spectral methods, etc.). Such methods have matured over the last 50 years and, in many cases, meet the robustness and computational efficiency standards required in practice. Our message here, as advocated in Section~\ref{sec:DT_models}, is that classical methods such as the Runge-Kutta time-stepping schemes can coexist in harmony with deep neural networks, and offer invaluable intuition in constructing structured predictive algorithms. Moreover, the implementation simplicity of the latter greatly favors rapid development and testing of new ideas, potentially opening the path for a new era in data-driven scientific computing. This will be further highlighted in the second part of this paper in which {\em physics informed neural networks} are put to the test of data-driven discovery of partial differential equations.\\

Finally, in terms of future work, one pressing question involves addressing the problem of quantifying the uncertainty associated with the neural network  predictions. Although this important element was naturally addressed in previous work employing Gaussian processes \cite{raissi2017numerical}, it not captured by the proposed methodology in its present form and requires further investigation.

\section*{Acknowledgements}
This work received support by the DARPA EQUiPS grant N66001-15-2-4055, the MURI/ARO grant W911NF-15-1-0562, and the AFOSR grant FA9550-17-1-0013. All data and codes used in this manuscript are publicly available on GitHub at \url{https://github.com/maziarraissi/PINNs}.

% \appendix
% \input{appendix.tex}


%% References


\bibliographystyle{model1-num-names}
\bibliography{sample.bib}






% \end{thebibliography}

\end{document}